\documentclass[letterpaper, 10 pt, conference]{ieeeconf}

\IEEEoverridecommandlockouts                              

\usepackage{amsmath,amsfonts}
\usepackage{algorithmic}
\usepackage{algorithm}
\usepackage{array}
\usepackage[caption=false,font=normalsize,labelfont=sf,textfont=sf]{subfig}
\usepackage{textcomp}
\usepackage{stfloats}
\usepackage{url}
\usepackage{verbatim}
\usepackage{graphicx}
\usepackage{cite}
\usepackage{booktabs}
\usepackage{multirow}
\usepackage{pifont}
\usepackage{xcolor}
\usepackage{bm}
\usepackage{tabularx}   
\usepackage{ragged2e}   
\usepackage{threeparttable}

\usepackage[colorlinks=true]{hyperref}
\hypersetup{
    colorlinks=true,
    urlcolor=magenta,
    linkcolor=black,
    citecolor=black
}

\newcolumntype{C}{>{\Centering\arraybackslash}X} 
\newcolumntype{L}{>{\RaggedRight\arraybackslash}X} 

\begin{document}

\title{
Learning Whole-Body Control for a Salamander Robot
}

\author{
Mengze Tian$^{1,\dagger}$,
Qiyuan Fu$^{1,\dagger,*}$,
Chuanfang Ning$^{1}$,
Javier Jia Jie Pey$^{1}$,
Auke Ijspeert$^{1}$%
\thanks{$^{1}$All authors are with the Biorobotics Laboratory (BIOROB), École polytechnique fédérale de Lausanne (EPFL), Lausanne 1015, Switzerland.}%
\thanks{$^{\dagger}$These authors contributed equally to this work.}
\thanks{$^{*}$Corresponding author: qiyuan.fu@epfl.ch.}
\thanks{This work has been submitted to the IEEE for possible publication. 
Copyright may be transferred without notice, after which this version may no longer be accessible.}
}

\maketitle
\thispagestyle{empty}
\pagestyle{empty}

\begin{abstract}
Amphibious legged robots inspired by salamanders are promising in applications in complex amphibious environments.
However, despite the significant success of training controllers that achieve diverse locomotion behaviors in conventional quadrupedal robots, most salamander robots relied on central-pattern-generator (CPG)-based and model-based coordination strategies for locomotion control. Learning unified joint-level whole-body control that reliably transfers from simulation to highly articulated physical salamander robots remains relatively underexplored. In addition, few legged robots have tried learning-based controllers in amphibious environments. 
In this work, we employ Reinforcement Learning to map proprioceptive observations and commanded velocities to joint-level actions, allowing coordinated locomotor behaviors to emerge.
To deploy these policies on hardware, we adopt a system-level real-to-sim matching and sim-to-real transfer strategy.
The learned controller achieves stable and coordinated walking on both flat and uneven terrains in the real world.
Beyond terrestrial locomotion, the framework enables transitions between walking and swimming in simulation, highlighting a phenomenon of interest for understanding locomotion across distinct physical modes.
A supplementary video is available at: \url{https://youtu.be/mMVRsBvLbYA}.
\end{abstract}


\section{Introduction}

Salamander robots are a promising amphibious platform for studying whole-body locomotion across both land and water~\cite{bizzi2000new,dickinson2000animals,grillner2006biological}. Their elongated body with an articulated spine and four limbs enables a single mechanical system to realize both limb-driven walking and body-driven swimming under distinct coordination patterns. This morphological versatility makes salamander-like robots a valuable testbed for investigating unified locomotion control within the same physical embodiment~\cite{IJSPEERT2008642,ijspeert2007swimming,ijspeert2014biorobotics}.

To coordinate the large number of degrees of freedom in locomotion, salamander robots are commonly controlled using biologically-inspired central pattern generator (CPG) controllers~\cite{IJSPEERT2008642,ijspeert2007swimming,ijspeert2014biorobotics,melo2023animal,thandiackal2021emergence,liu2025artificial}.
These approaches coordinate axial body undulation and limb stepping through a network of neurons that exhibit oscillatory signals, enabling rhythmic walking and swimming behaviors.
However, enabling adaptive and omnidirectional locomotion on a highly articulated salamander robot remains an open challenge.
In these systems, locomotion emerges from coupled whole-body dynamics involving axial bending, limb stepping, and environmental contact on both the feet and the body ~\cite{arreguit2025farms,hosseini2025minimalistic}.
Explicitly designing model-based controllers becomes increasingly challenging as contact interactions and oscillator network complexity grow rapidly with the body's number of degrees of freedom. This scalability issue limits their ability to achieve adaptive and omnidirectional locomotion~\cite{IJSPEERT2008642,ijspeert2007swimming,ijspeert2014biorobotics,horvat2015inverse,horvat2017spine,horvat2017model}, motivating learning-based approaches that can acquire whole-body control without model-based design.

Reinforcement Learning (RL) provides a natural framework for realizing this idea in practice~\cite{hwangbo2019learning,peng2018deepmimic,kaufmann2023champion}, but its potential for amphibious locomotion on salamander robots has not been fully explored.
Such locomotion is of significant biological interest~\cite{IJSPEERT2008642,ijspeert2007swimming,ijspeert2014biorobotics} and poses control challenges~\cite{shin2024fast,branicky2005introduction,ly2012learning}, as transitions between land and water induce shifts in propulsion mechanisms and contact conditions, leading to qualitatively different system dynamics.
From a systems perspective, such transitions give rise to distinct dynamical modes, effectively forming a hybrid dynamical system~\cite{branicky2005introduction}.
For robots with similar highly articulated morphologies, previous works rely on controllers such as trajectory generators or CPG for joint-level control~\cite{liu2025integrating,liu2025learning}, where the policy adjusts oscillator or phase parameters only to coordinate the predefined locomotion patterns instead of directly commanding all joints.
While this reduces control complexity, the policy modulates only a small set of parameters, limiting the range of achievable locomotion behaviors.
Learning a simple low-level joint-space policy that directly generates whole-body control and transferring it to a physical salamander robot remain less explored.

Realizing such learned whole-body control on a physical salamander robot further raises system-level considerations.
Discrepancies may arise across sensing, actuation, and kinematic layers.
At the sensing level, proprioceptive measurements are subject to noise, which affects the system’s state estimation~\cite{SORENSON1966219}.
At the actuation level, servo motors exhibit strongly nonlinear torque–angle characteristics, whose effective parameters are not directly known in simulation and therefore require identification from real hardware data~\cite{shi2025toddlerbot}.
At the kinematic level, mechanical backlash and compliance introduce kinematic deviations that alter the body dynamics.
Due to the serial-body morphology, such discrepancies accumulate along the kinematic chain and degrade whole-body coordination in sim-to-real transfer.
As a result, small discrepancies in sensing, actuation, and kinematics can degrade overall performance.

In this work, we study learning-based whole-body control for a salamander robot as a physical embodiment.
This work makes the following contributions:
\begin{itemize}
    \item An RL framework that learns joint-level whole-body control for terrestrial locomotion on a highly articulated salamander robot, where the policy directly outputs commands for all actuated joints rather than adjusting oscillator or phase parameters to generate motion.
    \item Development of a system-level real-to-sim matching and sim-to-real transfer pipeline, enabling learned whole-body locomotion policies to be deployed on hardware and achieve natural walking across both flat and uneven terrains in the real world.
    \item Beyond terrestrial locomotion, we formulate a transition across terrestrial and aquatic environments as a hybrid dynamical system and integrate it into the RL framework. We further validate in simulation that a unified policy can transition from land to water.

\end{itemize}

\begin{figure*}[t]
    \centering
    \includegraphics[width=\linewidth]{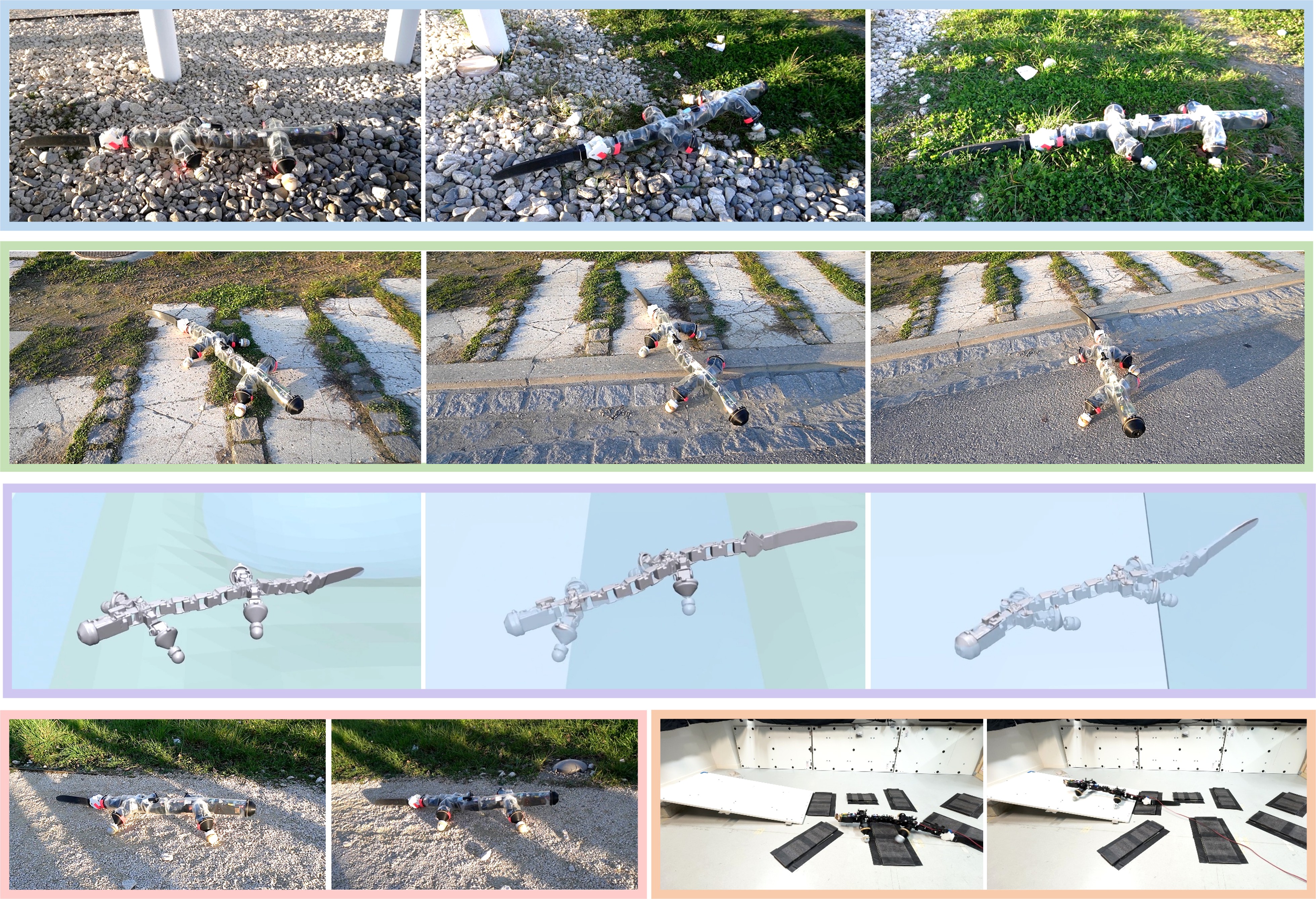}
    \caption{\textbf{Overview of the learned salamander locomotion across diverse terrains.}
    \textbf{Top row:} traversal from cobblestones to grass.
    \textbf{Second row:} traversal from stepped terrain, descending a height drop to flat ground.
    \textbf{Third row:} transition from land to water in simulation.
    \textbf{Bottom row:} locomotion on slippery sand (left) and rough terrain in the lab (right).}
    \label{fig:demo_}
\end{figure*}

\section{Background and Setting}

\subsection{Locomotion in Salamander-like Robots}

Previous salamander robots were primarily controlled by CPGs, which coordinate rhythmic body movements for both terrestrial and aquatic locomotion~\cite{IJSPEERT2008642,ijspeert2007swimming,ijspeert2014biorobotics}. CPGs regulate the oscillatory patterns that enable walking and swimming by coordinating axial and limb motions.
This neural mechanism modulates movement through a few simple control parameters, providing a biological model for locomotion.
While CPGs offer a biologically inspired approach, RL has become a powerful tool for learning complex behaviors through environmental interaction~\cite{hwangbo2019learning,peng2018deepmimic,kaufmann2023champion,liu2025learning}.

\subsection{Hybrid Dynamical System}
\label{hybrid_dynamics}

Hybrid dynamical systems describe processes that combine continuous dynamics with discrete mode transitions, where each mode is governed by a different set of dynamics~\cite{goebel2009hybrid}. 
Amphibious locomotion naturally exhibits such hybrid characteristics. 
On land, motion is dominated by contact-driven rigid-body dynamics that provide support and propulsion, whereas in water, contact forces vanish and motion is governed primarily by buoyancy and hydrodynamic drag. 
These two interaction regimes therefore correspond to distinct continuous dynamical modes within a unified physical system.

For terrestrial locomotion, we rely on MuJoCo’s standard rigid-body contact dynamics~\cite{todorov2012mujoco}. 
To model aquatic locomotion, we adopt a simplified hydrodynamic approximation that captures the dominant fluid forces while remaining computationally efficient and fully compatible with MuJoCo’s rigid-body pipeline. 
Specifically, buoyancy and drag forces are computed directly from each body segment’s instantaneous kinematics.

\textbf{Buoyancy:}
For segments below the water surface, an upward buoyancy force is applied:
\begin{equation}
\mathbf{F}_{b}
= k_b \, m g \, \mathbf{e}_z
- k_d \, \dot{z} \, \mathbf{e}_z ,
\end{equation}
where $\mathbf{e}_z$ denotes immersed depth, $m$ is the segment mass, $g$ is gravitational acceleration, $\dot{z}$ is the vertical velocity, $k_b, k_d$ denote buoyancy and damping coefficients.

\textbf{Hydrodynamic Drag:}
The hydrodynamic drag does not assume a strictly laminar or turbulent regime. Instead, it adopts a combined formulation that captures both linear and quadratic forces.
Drag forces and torques are computed for each submerged body segment from its velocity relative to the surrounding fluid.
In the local body frame, translational and rotational drag are approximated as linear and quadratic functions of the corresponding velocities,
\begin{align}
    \mathbf{F}_{d}
    &= -\mathbf{C}_v^{\text{lin}} \, \mathbf{v}
       -\mathbf{C}_v^{\text{quad}} \bigl(\|\mathbf{v}\|\odot \mathbf{v}\bigr), \\
    \boldsymbol{\tau}_{d}
    &= -\mathbf{C}_\omega^{\text{lin}} \, \boldsymbol{\omega}
       -\mathbf{C}_\omega^{\text{quad}} \bigl(\|\boldsymbol{\omega}\|\odot \boldsymbol{\omega}\bigr),
\end{align}
where $\mathbf{v}, \boldsymbol{\omega} \in \mathbb{R}^3$ are the linear and angular velocities of the segment in its local frame,
$\mathbf{C}_v^{\text{lin}}, \mathbf{C}_v^{\text{quad}}, \mathbf{C}_\omega^{\text{lin}}, \mathbf{C}_\omega^{\text{quad}}$ are diagonal gain matrices that absorb the effects of geometry, viscosity, and fluid density, and $\odot$ denotes element-wise multiplication.
The drag gain matrices
$\mathbf{C}_v^{\text{lin}},
 \mathbf{C}_v^{\text{quad}},
 \mathbf{C}_\omega^{\text{lin}},
 \mathbf{C}_\omega^{\text{quad}}$ are computed analytically from each body segment’s geometry and physical properties~\cite{todorov2012mujoco}.
The resulting forces and torques are then transformed back to the world frame and applied as external loads.

Together with contact-driven rigid-body dynamics on land, this model defines the two continuous dynamical modes that constitute the hybrid locomotion system studied in this work.

\subsection{Physical Platform}

In this work, we use a salamander-inspired amphibious robot as the physical embodiment for studying whole-body locomotion across terrestrial environments.
The robot features a highly articulated serial body with multiple axial and limb degrees of freedom, actuated by position-controlled Dynamixel XM430 servo motors.
It is equipped with an onboard 9-DoF inertial measurement unit (IMU) based on the LSM6DSOX and LIS3MDL sensors, and a Raspberry Pi Zero 2W for onboard control and data processing.

To enable real-time execution of the learned policies on this embedded platform, we deploy the trained models using the ONNX runtime library~\cite{onnx}.
The policies are exported to ONNX format and executed on the ARM architecture of the Raspberry Pi Zero 2W.
A ROS2-based control framework integrates state estimation, policy inference, and low-level motor command publishing onboard without external computation. The operator's laptop only sends gamepad input for goal velocities over Wi-Fi.
The robot is powered by a DC power supply at 14 V during walking tests or a 3-cell Lithium-Polymer battery during outdoor tests.

\section{Method}
\subsection{Action Space}

We formulate whole-body locomotion as a learning-based joint-space control problem.
At each control step $t$, the policy outputs an incremental joint command
\begin{equation}
\mathbf{a}_t = \Delta \mathbf{q}_{\mathrm{des}} \in \mathbb{R}^{n_q},
\end{equation}
where $\Delta \mathbf{q}_{\mathrm{des}}$ denotes the desired joint angle offset,
and $n_q$ is the number of actuated joints of the salamander robot.

Instead of directly predicting absolute joint angles, the action is defined as a residual offset added to a fixed nominal joint angle $\mathbf{q}_{\mathrm{nominal}} \in \mathbb{R}^{n_q}$,
\begin{equation}
\mathbf{q}_{\mathrm{des}} = \mathbf{q}_{\mathrm{nominal}} + \Delta \mathbf{q}_{\mathrm{des}},
\end{equation}
where $\mathbf{q}_{\mathrm{des}} \in \mathbb{R}^{n_q}$ denotes the desired joint angle.
This residual action formulation anchors the learned policy around a physically meaningful posture prior, facilitating stable exploration during training~\cite{silver2018residual,johannink2019residual,carvalho2022residual}.
The desired joint angles $\mathbf{q}_{\mathrm{des}}$ are tracked by a low-level joint-space PD controller, which produces the corresponding joint torques $\boldsymbol{\tau} \in \mathbb{R}^{n_q}$.

\begin{figure*}[t]
    \centering
    \includegraphics[width=\textwidth]{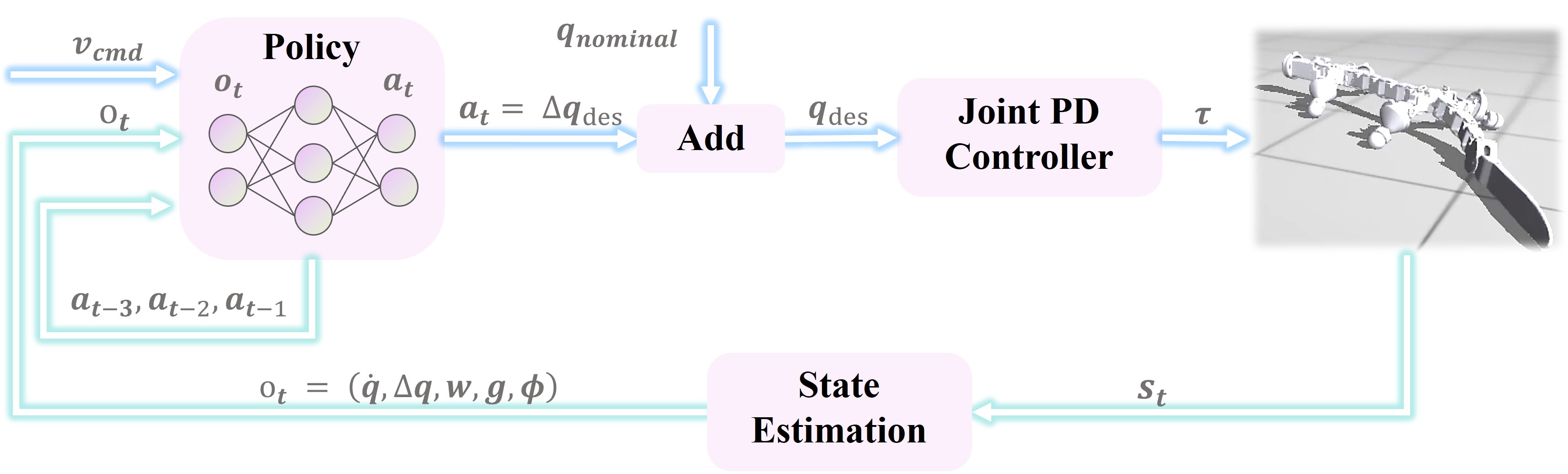}
    \caption{Control architecture of the learned joint-space controller.
    The policy receives commanded planar velocity $\mathbf{v}_{\mathrm{cmd}}$ and a proprioceptive observation
    $\mathbf{o}_t = (\mathbf{q}, \Delta\mathbf{q}, \boldsymbol{\omega}, \mathbf{g}, \boldsymbol{\phi})$,
    together with a short history of past actions.
    The policy outputs joint angle residuals $\Delta \mathbf{q}_{\mathrm{des}}$, which are added to a nominal joint angle $\mathbf{q}_{\mathrm{nominal}}$ to form the desired joint angle $\mathbf{q}_{\mathrm{des}}$.
    A low-level joint-space PD controller tracks $\mathbf{q}_{\mathrm{des}}$ and produces joint torques $\boldsymbol{\tau}$.
    }

    \label{fig:architecture}
\end{figure*}

\subsection{Observation Space}

The policy receives a proprioceptive observation $\mathbf{o}_t$ obtained from onboard state estimation.
As illustrated in Fig.~\ref{fig:architecture}, the observation is defined as
\begin{equation}
\mathbf{o}_t =
\bigl(
\dot{\mathbf{q}},
\Delta \mathbf{q},
\boldsymbol{\omega},
\mathbf{g},
\boldsymbol{\phi}
\bigr),
\end{equation}
where $\dot{\mathbf{q}}$ denotes the joint velocities, $\Delta \mathbf{q}$ denotes joint angle differences,
$\boldsymbol{\omega}$ is the angular velocity measured by the IMU, $\mathbf{g}$ is the gravity vector expressed in the body frame measured by the IMU,
and $\boldsymbol{\phi}$ represents the current gait phase.

\textbf{Phase input:}
To facilitate foot clearance without introducing an explicit gait generator,
we define a phase input $\boldsymbol{\phi}$~\cite{siekmann2021sim,margolis2023walk,liao2025berkeley}. 
It is provided as an additional input to the policy and is used in computing phase-conditioned reward terms that encourage foot clearance, as described in Sect.~\ref{sec:reward}. 
Importantly, the phase variable does not generate joint commands and is not tracked by the low-level controller. 
All joint commands are directly produced by the learned policy.

\textbf{Observation without linear velocity:}
The IMU linear velocity is excluded from the observation.
For the salamander robot, estimating linear velocity requires inertial integration, which is prone to drift and difficult to stabilize due to the robot's compact body and limited sensing capabilities.
To compensate for the absence of explicit linear velocity information, the policy additionally takes a short history of past actions
$\{\mathbf{a}_{t-1}, \mathbf{a}_{t-2}, \mathbf{a}_{t-3}\}$ as a separate input,
which provides temporal information for locomotion dynamics.

\subsection{Reward Functions}
\label{sec:reward}
The reward function is designed to promote stable and efficient whole-body locomotion while tracking commanded motion.
In particular, we define the tracking rewards using the equivalent center-of-mass (CoM) linear and angular velocities, rather than velocities measured at local sensors by the IMU.
Due to the oscillatory motions of the body during salamander locomotion, velocities measured by IMU exhibit periodic fluctuations and do not reliably reflect the overall motion of the system.
In contrast, CoM velocities reduce the influence of oscillations and better capture the overall motion.

At each timestep $t$, the total reward is defined as a weighted sum of four components,
\begin{equation}
r_t =
w_v\, r_v
+ w_\omega\, r_\omega
+ w_{\text{energy}}\, r_{\text{energy}}
+ w_{\text{phase}}\, r_{\text{phase}} .
\end{equation}
Individual reward terms are summarized in Tab.~\ref{tab:reward_functions}.

The linear and angular velocity tracking terms, $r_v$ and $r_\omega$, constitute the primary task objectives and encourage the robot to follow the commanded planar velocity and yaw rate.
Energy consumption is penalized through the instantaneous mechanical power at each joint,
$r_{\text{energy}}$, promoting efficient whole-body coordination.
The foot phase consistency term $r_{\text{phase}}$ regularizes foot height by comparing each foot's vertical height $z^{\text{foot}}_j$ to a smooth phase-conditioned reference profile $z^{\text{ref}}_j(\phi_j)$ parameterized by a Bézier curve~\cite{liao2025berkeley}. Here, $\phi_j \in [-\pi,\pi]$ denotes the gait phase of foot $j$. This formulation introduces a weak rhythmic prior without explicitly enforcing a fixed gait.
In the experiments, we set $w_v = 1.0dt$, $w_\omega = 0.5dt$, $w_{\text{energy}} = 1\times10^{-3}dt$, and $w_{\text{phase}} = 1.0dt$ with $dt = 0.02$, where all physical quantities are expressed in standard SI units.

\begin{table}[H]
\centering
\caption{Reward Functions}
\begin{tabular}{l c}
\toprule
\textbf{Reward Term} & \textbf{Expression} \\
\midrule
Linear Velocity Tracking
& $
r_v = \exp\!\left(
-\frac{\left\lVert
\mathbf{v}_{\mathrm{cmd},xy}
-
\mathbf{v}_{b,xy}
\right\rVert^2}{\sigma_v}
\right)
$ \\[6pt]

Angular Velocity Tracking
& $
r_\omega = \exp\!\left(
-\frac{
\left(
\omega_{\mathrm{cmd},z}
-
\omega_{b,z}
\right)^2}{\sigma_\omega}
\right)
$ \\[6pt]

Energy Cost
& $
r_{\text{energy}}
=
- \sum_i
\big|
\dot{q}_i\, \tau_i
\big|
$ \\[6pt]

Feet Phase Consistency
& $
r_{\text{phase}}
=
\exp\!\left(
-\frac{
\sum_j
\left(
z^{\text{foot}}_j
-
z^{\text{ref}}_j(\phi_j)
\right)^2}{\sigma_{\text{phase}}}
\right)
$ \\[4pt]

\bottomrule
\end{tabular}
\label{tab:reward_functions}
\end{table}

\subsection{Training Details}

\begin{figure}[!h]
    \centering
    \includegraphics[width=\linewidth]{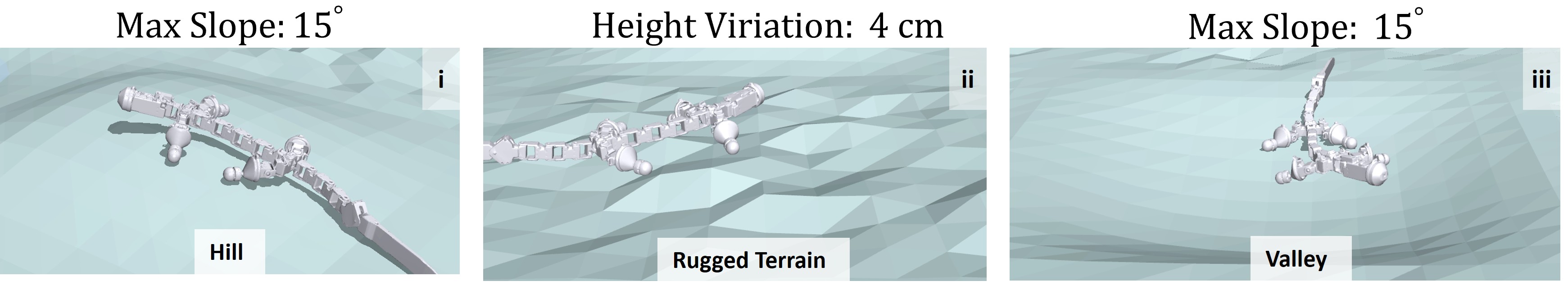}
    \caption{Training terrains used for locomotion learning with the robot 
    having a maximum ground clearance of $10.2\,\mathrm{cm}$ when all tibia links are vertical.
    (i) Hill terrain with the max slope angle of $15^\circ$. 
    (ii) Rugged terrain with the max height variation of $4\,\mathrm{cm}$. 
    (iii) Valley terrain with the max slope angle of $15^\circ$.}
    \label{fig:training_env}
\end{figure}

All policies are trained in simulation using the MuJoCo Playground framework~\cite{zakka2025mujoco}, as visualized in Fig.~\ref{fig:training_env}.
Training is performed with PPO~\cite{schulman2017proximal} on a single NVIDIA RTX 4070 GPU for approximately 1.5 hours, corresponding to $3\times10^8$ steps, as summarized in Tab.~\ref{tab:ppo_config}.
The policy operates at a control frequency of 50~Hz, while desired joint positions are tracked by a low-level joint PD controller with gains $K_p = 15~\mathrm{Nm/rad}$ and $K_d = 0.15~\mathrm{Nms/rad}$.

\begin{table}[!h]
\centering
\caption{PPO Training Configuration}
\begin{tabular}{l c}
\toprule
\textbf{Parameter} & \textbf{Value} \\
\midrule

Training steps & $3\times10^{8}$ \\

Number of environments & 8192 \\

Episode length & 1000 \\

Discount factor $\gamma$ & 0.97 \\

Learning rate & $3\times10^{-4}$ \\

Entropy coefficient & $1\times10^{-2}$ \\

Unroll length & 20 \\

Minibatches per update & 32 \\

Updates per batch & 4 \\

Batch size & 256 \\

Gradient clipping & 1.0 \\

Observation normalization & Enabled \\

Policy network & MLP (512, 256, 128) \\

Value network & MLP (512, 256, 128) \\

\bottomrule
\end{tabular}
\label{tab:ppo_config}
\end{table}

\subsection{Sim-to-Real Transfer}

This work addresses sim-to-real transfer for a highly serial, servo-actuated salamander robot, where modeling errors of the mechanical properties and motor dynamics accumulate along the kinematic chain and result in significant differences on whole-body dynamics.
To bridge this gap, we implement a full-stack alignment pipeline spanning the observation, action, and kinematic layers. 
The objective is to ensure consistency between (i) the observations provided to the policy, (ii) the action commands admissible by the hardware, and (iii) the resulting kinematics observed on the physical system.

\textbf{Observation Alignment:}
To reduce discrepancies between simulated and real proprioceptive inputs, we inject observation noise during training based on observation measurements collected on the physical robot. The real sensor data are recorded while the robot executes reference motions generated from previous studies. From these recordings, we determine bounded noise ranges such that the injected noise envelopes the variability observed on hardware.

\textbf{Action Alignment:}
To ensure that policy outputs remain within the tracking capabilities of the actuators, we apply a first-order low-pass filter to the commanded joint position offsets before execution~\cite{li2021reinforcement}. In addition, a servo envelope model is used to capture actuator saturation, velocity-dependent torque limits, and braking asymmetry~\cite{shi2025toddlerbot}.
The motor parameters in simulation are further calibrated using tracking data collected from the physical robot, improving the consistency between simulated and real actuator responses.

\textbf{Kinematic Alignment:}
Mechanical clearances in the real robot introduce backlash that accumulates under oscillatory locomotion, reducing reachable leg workspace and lowering foot height. To account for this, we augment the simulation model with two auxiliary rotational DoFs at the front and back girdle connections, each allowing $4^\circ$ passive rotation.
The magnitude of passive rotation is determined by comparing simulated and real kinematic data, and the simulated model reproduces the accumulated orientation drift observed on hardware, yielding foot trajectories that more closely match those of the physical robot.

\begin{figure}[!h]
    \centering
    \includegraphics[width=\linewidth]{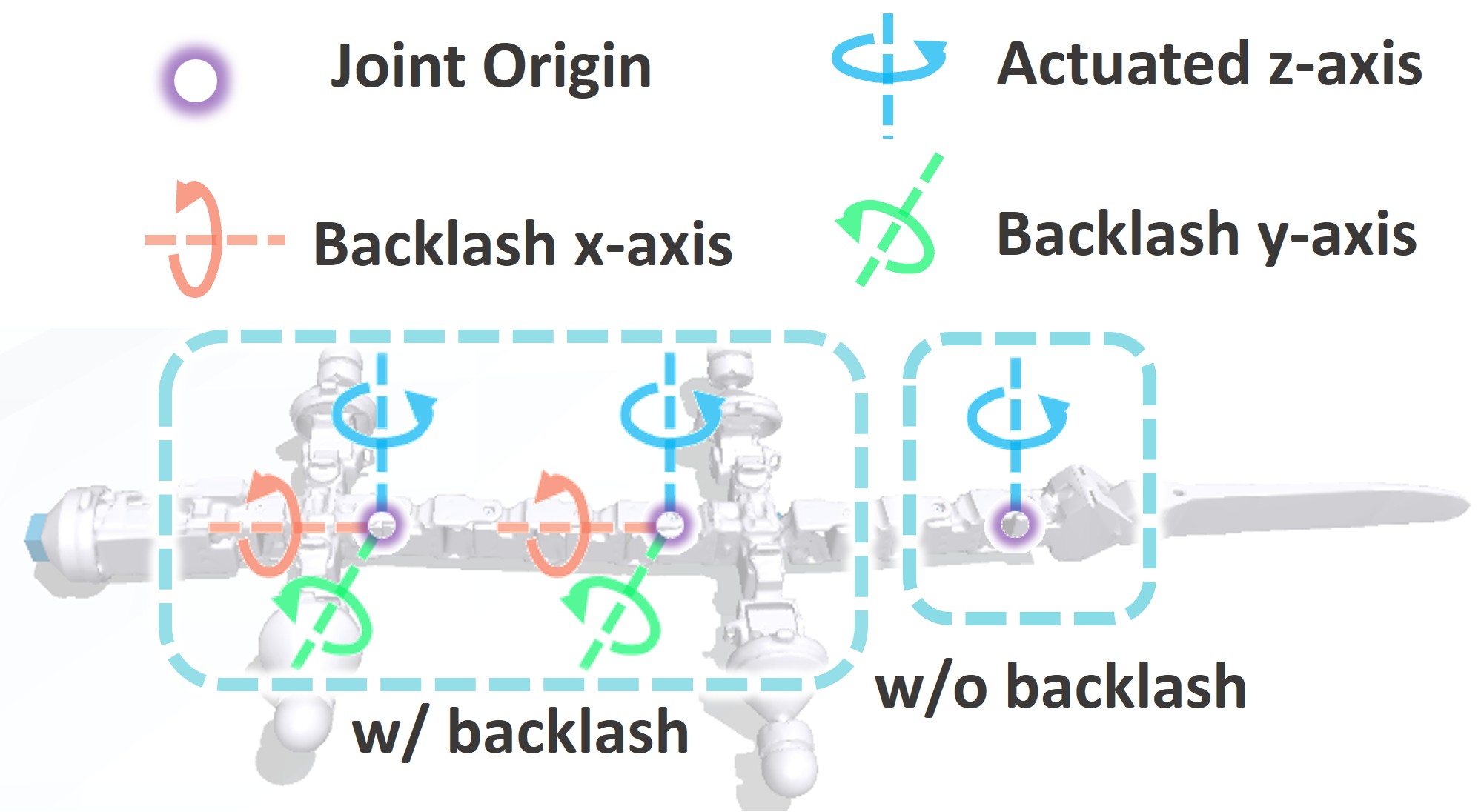}
    \caption{Kinematic modeling of structural backlash.
    In addition to the actuated joint rotation about the $z$-axis (blue), two auxiliary passive rotational degrees of freedom about the $x$-axis and $y$-axis (orange and green) are introduced at the front and back girdle connections to model accumulated mechanical clearances.
    All other joints in the model remain rigid and do not include additional passive degrees of freedom.}

    \label{fig:backlash_mechanism}
\end{figure}

To improve robustness to unmodeled dynamics and environmental variability, we expose the policy to external disturbances by applying randomized forces to the torso during training.
In addition, domain randomization is applied over a subset of physical and control parameters~\cite{peng2018sim}, including contact friction, mass distribution, joint friction losses, and PD gains. The training environment is visualized in Fig.~\ref{fig:training_env}.










\subsection{Transition across Land and Water}
\label{transition_formulation}
\begin{figure}[!h]
    \centering
    \includegraphics[width=\linewidth]{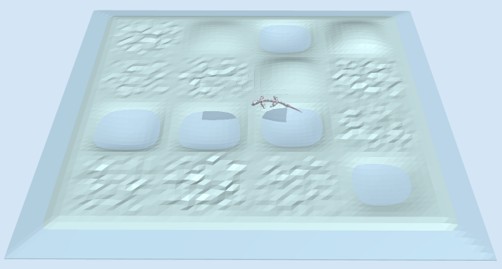}
    \caption{
    Training environment for the transition task. Blue regions indicate water areas. 
    }
    \label{fig:transition_env}
\end{figure}

As described in Sect.~\ref{hybrid_dynamics}, terrestrial and aquatic locomotion is defined by different dominant physical interactions in MuJoCo. 
On land, motion follows contact-based rigid-body dynamics, whereas in water, buoyancy and hydrodynamic drag are additionally applied to each body segment.

We formulate the resulting environment-dependent dynamics as
\begin{equation}
    \mathbf{s}_{t+1} = f^{\sigma}(\mathbf{s}_t, \mathbf{a}_t),
\end{equation}
where $\sigma \in \{\text{land}, \text{water}\}$ indicates the locomotion mode. 
Here, $f^{\sigma}$ reflects the different physical simulation configurations defined in MuJoCo and serves to characterize the hybrid dynamical system; it is not used for explicit model-based control or switching.

A single unified policy can be trained across both modes, as illustrated in Fig.~\ref{fig:transition_env}.
To allow the policy to condition on the environment, a binary mode indicator $\sigma$ is appended to the proprioceptive observation:
\begin{equation}
    \mathbf{o}_t =
    \big(
    \dot{\mathbf{q}},\,
    \Delta \mathbf{q},\,
    \boldsymbol{\omega},\,
    \mathbf{g},\,
    \boldsymbol{\phi},\,
    \sigma
    \big).
\label{transition_obs}
\end{equation}

\section{Results}

\subsection{Locomotion Performances}

\begin{figure*}[t]
    \centering
    \includegraphics[width=\linewidth]{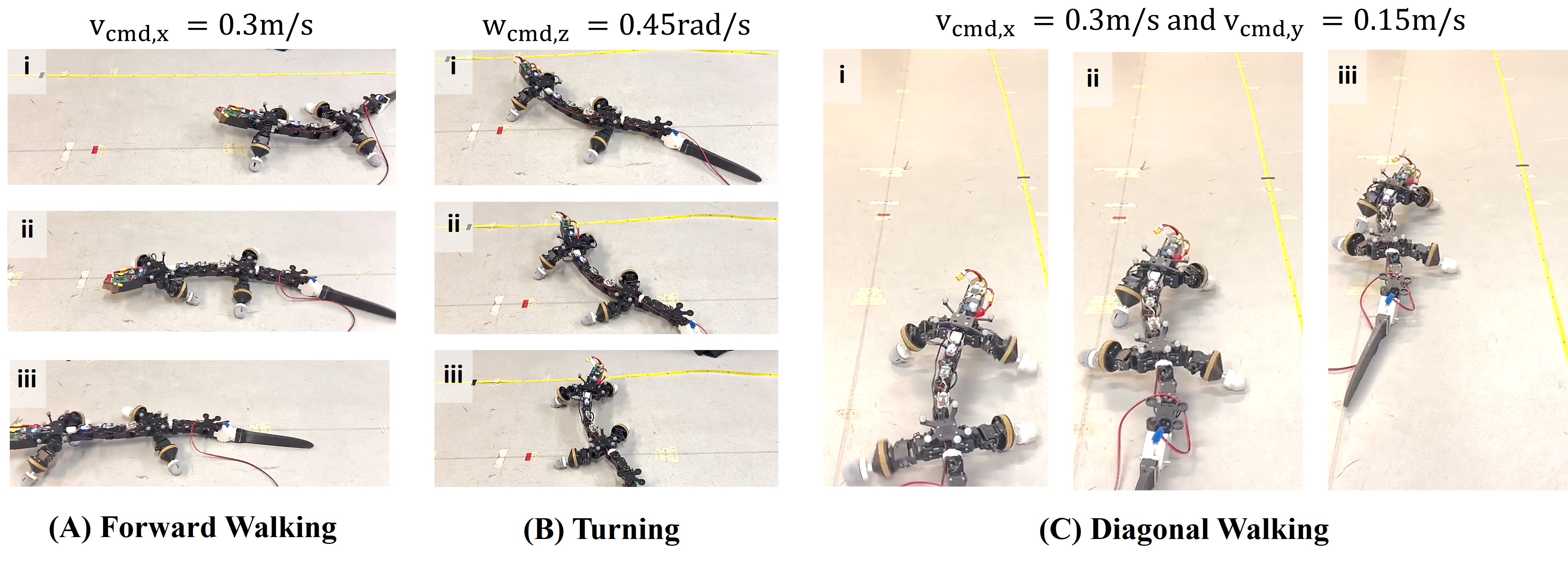}
    \caption{
    Real-world omnidirectional locomotion behaviors.
    (A) Forward walking under a constant forward velocity command, exhibiting coordinated axial–limb propulsion.
    (B) Turning behavior driven by a nonzero yaw-rate command, where asymmetric limb coordination induces rotational motion.
    (C) Diagonal walking resulting from simultaneous forward and lateral velocity commands, demonstrating continuous adaptation of whole-body coordination to mixed-direction motion.
    }

    \label{fig:realworld}
\end{figure*}

\textbf{Omnidirectional Walking:} We evaluate the learned controller on omnidirectional walking on flat ground.
The robot achieves stable omnidirectional walking under diverse commanded velocities.
As shown in Fig.~\ref{fig:realworld}, the policy generates coordinated motions that adapt to different commands. A forward velocity command leads to consistent rhythmic walking, a nonzero yaw rate produces smooth turning through gait asymmetry, and combined forward and lateral commands result in motion along diagonal directions.
These results show that a single policy enables omnidirectional and flexible locomotion under varying velocity commands.

\begin{figure}[!h]
    \centering
    \includegraphics[width=\linewidth]{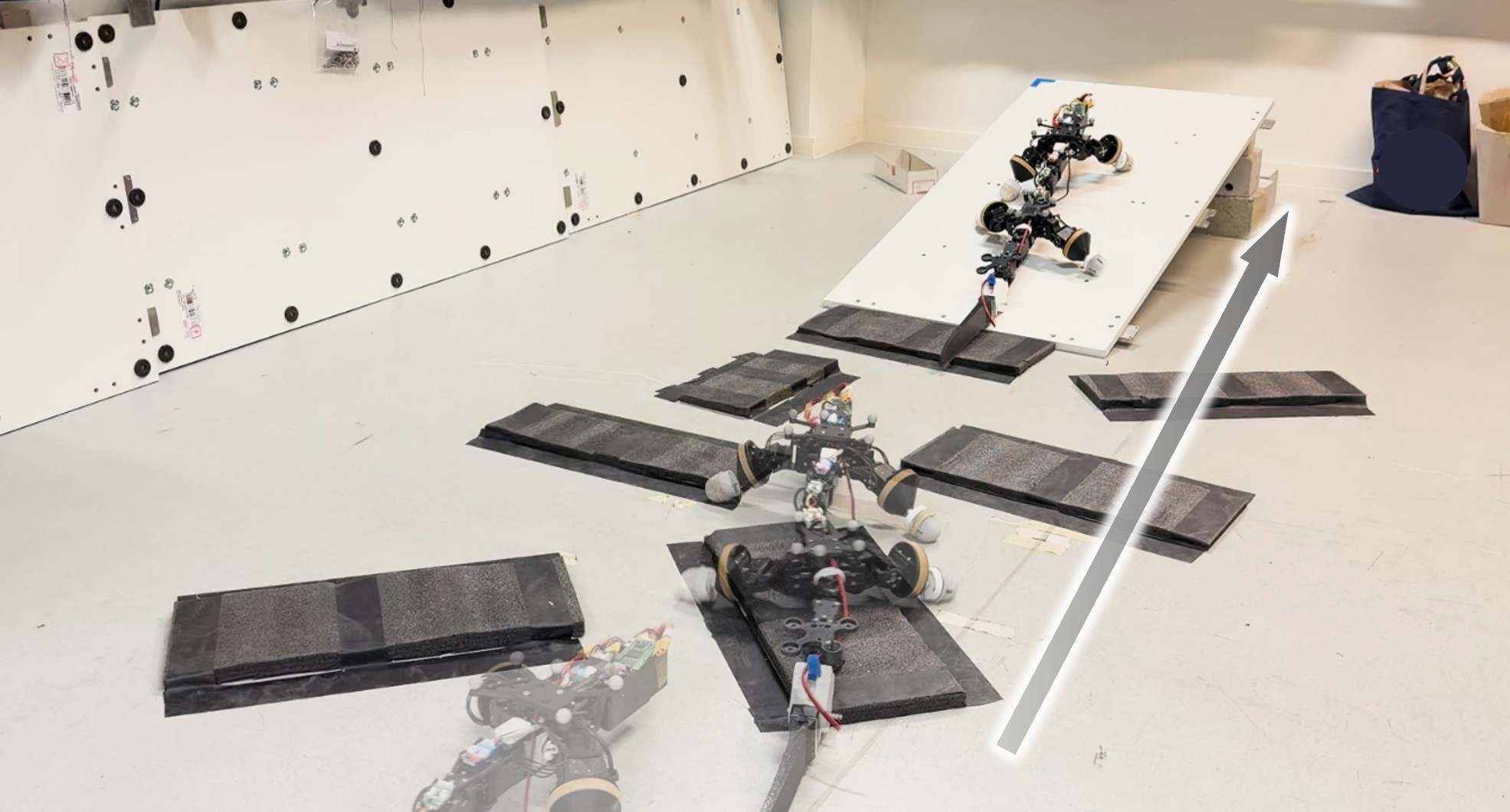}
    \caption{
    The robot successfully traverses rough-medium terrain. The terrain consists of inclined ramps and fragmented surface patches with varying elevations. The arrow indicates the direction of motion over time.
    }

    \label{fig:real_world_rough_lab}
\end{figure}

\textbf{Walking on Rough Terrain:} We evaluate locomotion over uneven terrain. Quantitative results are summarized in Table~IV. We visualize the Rough-Medium setting reported in Table~IV, where the robot traverses ramps and fragmented elevated patches, as shown in Fig.~\ref{fig:real_world_rough_lab}.
We measure the forward velocity $v_{b,x}$ in the body frame by dividing the front girdle displacement by the total traversal time.
The achieved forward velocity decreases with terrain difficulty, and the robot sustains locomotion across all tested conditions.
On flat terrain with a fixed forward command of 0.3~$m/s$, the average forward velocity is 0.23~$m/s$. 
On rough terrains under manual joystick control, velocities remain in a comparable range with degradation as ruggedness and slope increase.
Overall, these results show that the learning framework supports different terrestrial locomotion, enabling walking on flat ground and adaptive traversal over rough terrain.

\begin{table}[!h]
\centering
\caption{Terrain definitions and corresponding real-world forward velocity statistics. Each result includes data aggregated from at least three trials. Clearance (\%) is defined as terrain ruggedness normalized by the robot's ground clearance.}
\label{tab:terrain_and_velocity}

\begin{tabular}{l c c c}
\toprule
\textbf{Terrain} & \textbf{Ruggedness (cm)} & \textbf{Clearance (\%)} & \textbf{Slope (°)} \\
\midrule
Flat & 0 & 0.0 & 0 \\
Rough--Easy & 2 & 19.6 & 15 \\
Rough--Medium & 4 & 39.2 & 15 \\
Rough--Hard & 4 & 39.2 & 25 \\
\bottomrule
\end{tabular}

\vspace{0.6em}

\begin{tabular}{l l c}
\toprule
\textbf{Terrain} & \textbf{Command Type} & $v_{b,x}$ (m/s) \\
\midrule
Flat & Fixed forward velocity (0.3 m/s) & 0.23 $\pm$ 0.01 \\
Rough--Easy & Manual joystick control (0--0.3 m/s) & 0.19 $\pm$ 0.03 \\
Rough--Medium & Manual joystick control (0--0.3 m/s) & 0.18 $\pm$ 0.01 \\
Rough--Hard & Manual joystick control (0--0.3 m/s) & 0.17 $\pm$ 0.04 \\
\bottomrule
\end{tabular}

\end{table}

\subsection{Evaluation of Sim-to-Real Transfer}

We evaluate sim-to-real transfer through system-level alignment analysis and a velocity tracking experiment using the learned controller.

\begin{figure}[!h]
    \centering
    \includegraphics[width=\linewidth]{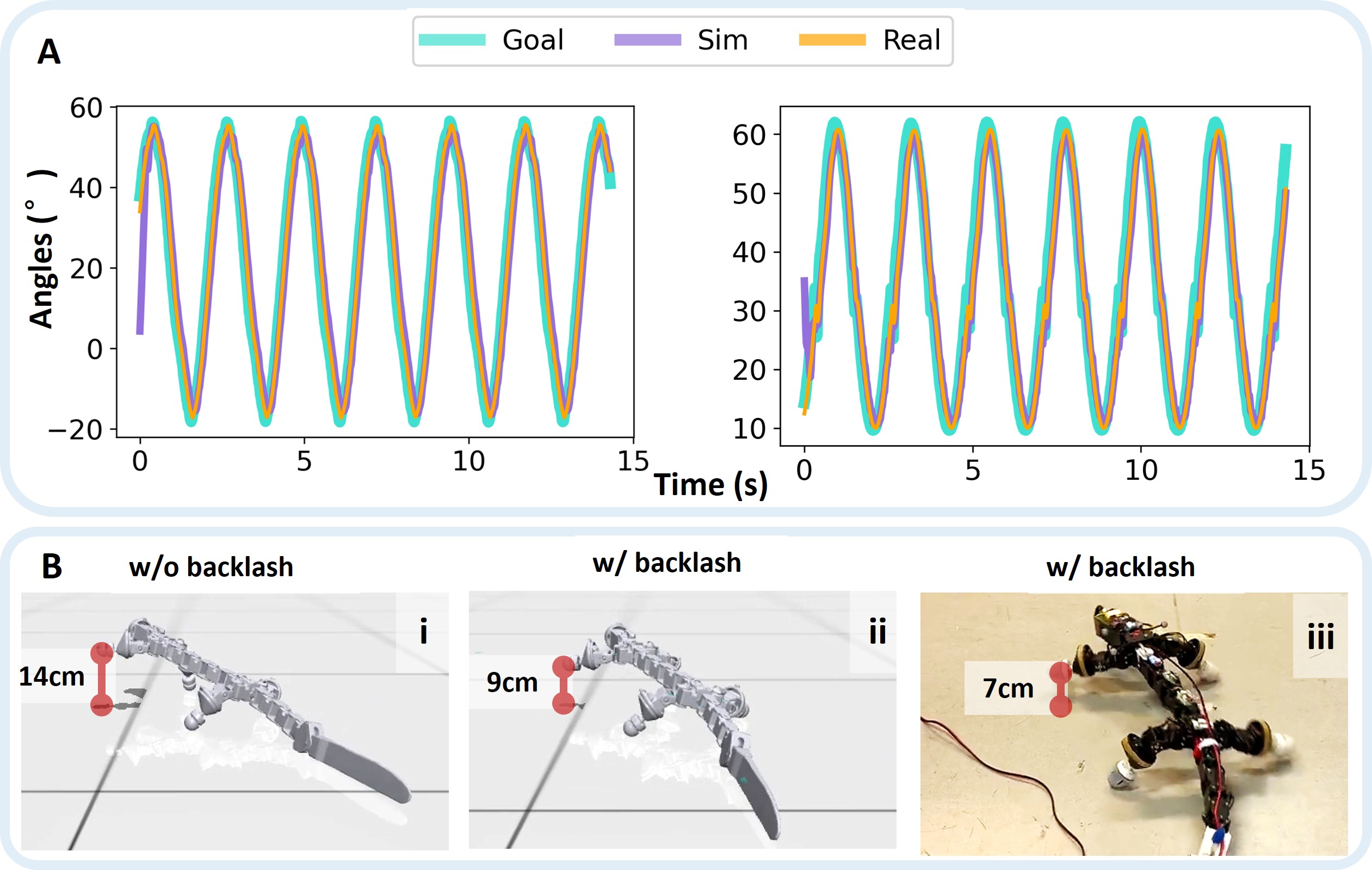}
    \caption{
    (A) Joint angle tracking of two representative joints on the front-left (FL) leg: the hip joint (left) and the foot joint (right). The cyan curve denotes the target command (Goal), the purple curve the simulated response (Sim), and the orange curve the real robot measurement (Real).
    (B) Visualization of modeling backlash.
    }

    \label{fig:sim2real_system}
\end{figure}

\textbf{Evaluation of System Alignment:} We evaluate sim-to-real consistency at the system level, including action alignment and kinematic alignment. As shown in Fig.~\ref{fig:sim2real_system}A, the calibrated simulated motors reproduce the joint angle trajectories of the physical robot, with simulated responses remaining close to real measurements under the same commands. 
We further examine the effect of structural backlash modeling. As illustrated in Fig.~\ref{fig:sim2real_system}B, without backlash the simulated front foot clearance is overestimated. Introducing passive rotational degrees of freedom at the torso--girdle connections reduces the predicted swing foot height and produces trajectories that more closely match those observed on hardware. These results show that incorporating backlash improves kinematic alignment between simulation and the real robot.

\begin{figure}[!h]
    \centering
    \includegraphics[width=\linewidth]{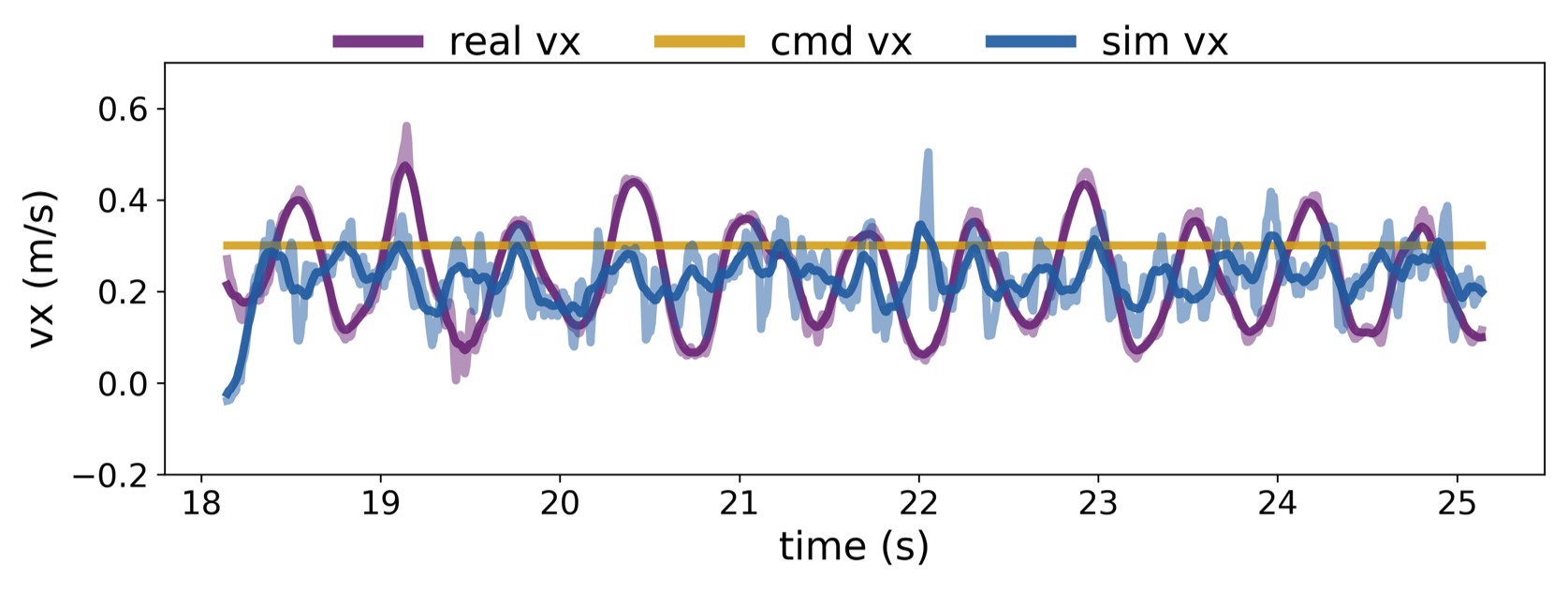}
    \caption{
    Sim-to-real tracking evaluation of the robot forward velocity. The yellow curve (cmd vx) denotes the commanded forward velocity, the blue curve (sim vx) represents the velocity in simulation, and the purple curve (real vx) shows the velocity measured on the real robot.
    }

    \label{fig:sim2real_tracking}
\end{figure}

\textbf{Evaluation of Tracking Performances:}
We evaluate forward velocity tracking, where the real-world speed of the front girdle is measured using an OptiTrack motion capture system.
As shown in Fig.~\ref{fig:sim2real_tracking}, both simulated and real forward velocities follow the commanded profile with similar temporal trends. Tracking errors are observed in both domains, and deviations from the command are present in the measured velocities. Despite these discrepancies, the overall evolution of forward velocity remains similar between simulation and hardware, showing that the policy produces comparable locomotion behavior on the physical platform.

\subsection{Transition from Land to Water}

\begin{figure}[!h]
    \centering
    \includegraphics[width=\linewidth]{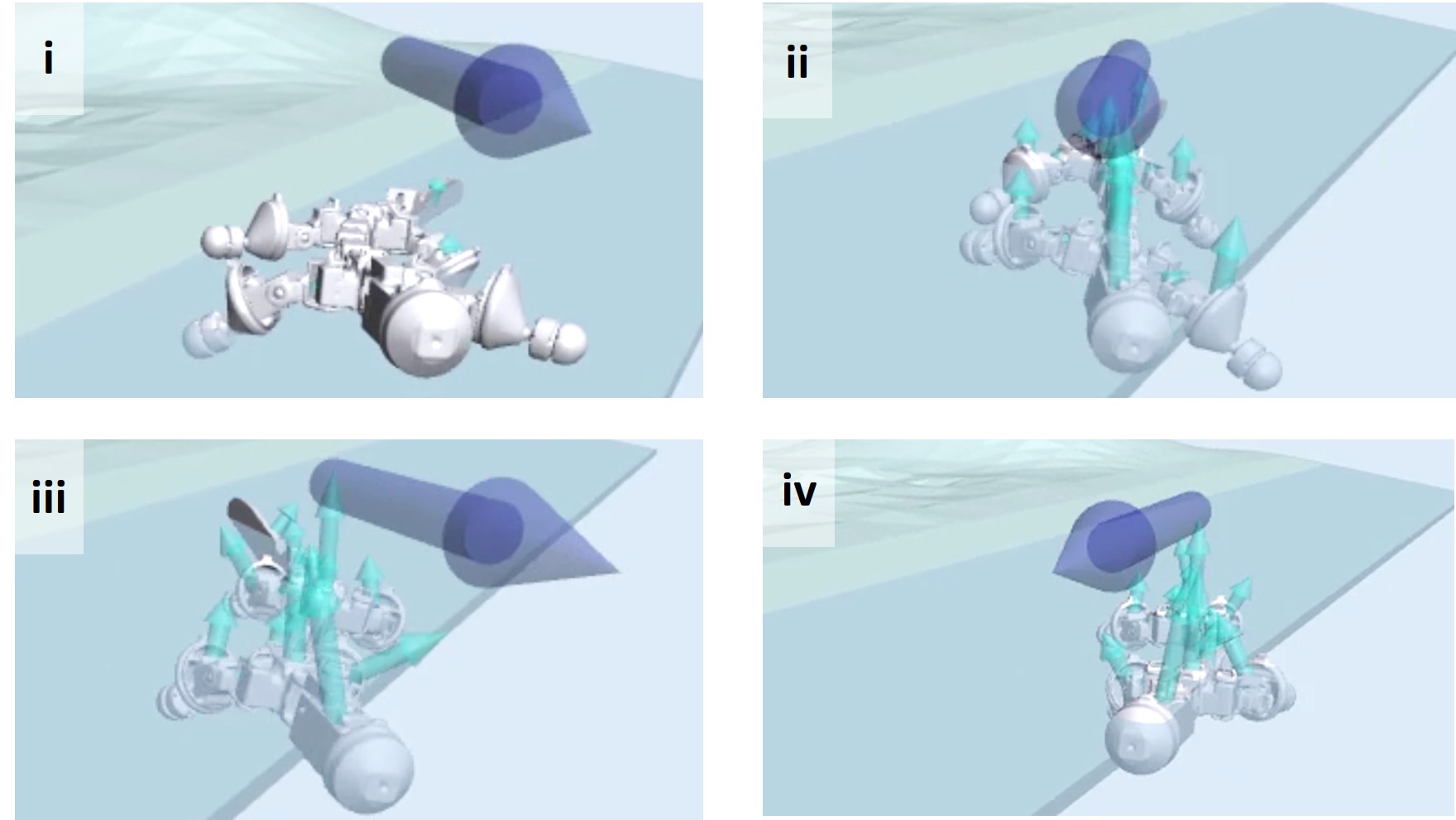}
    \caption{
    Snapshots of the robot during the transition from land locomotion to swimming. The sequence (i–iv) illustrates the robot approaching the water, entering the water region, and progressively shifting from limb-driven walking to body-driven undulatory swimming.
    }

    \label{fig:transition_snapshots}
\end{figure}

\begin{figure}[!h]
    \centering
    \includegraphics[width=\linewidth]{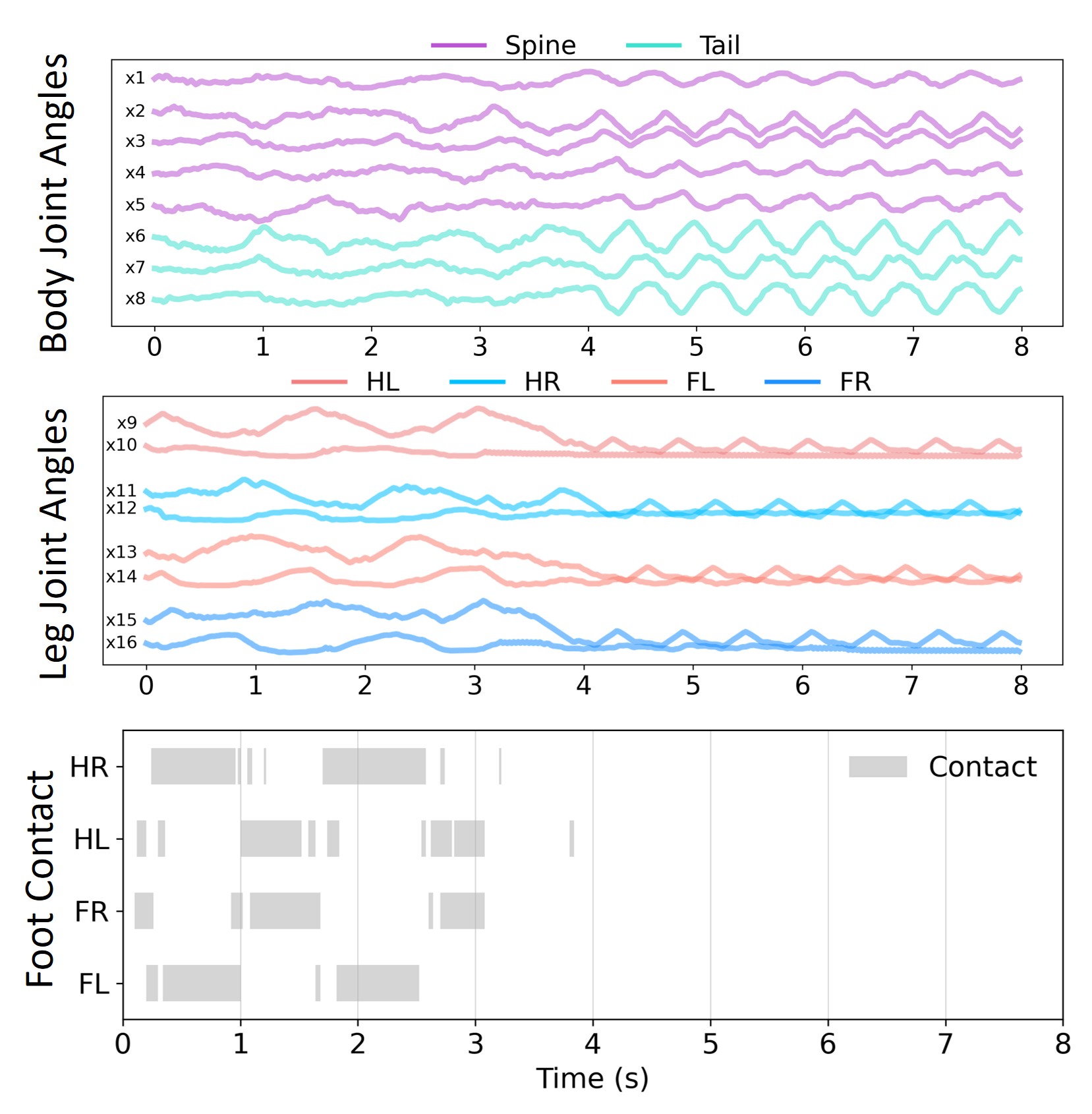}
    \caption{
    Transition from terrestrial to aquatic locomotion in simulation, where $x_i$ represents the joint angle.
    (Top) Body joint trajectories for the spine and tail joints. 
    (Middle) Leg joint trajectories for the four limbs (HL, HR, FL, FR). 
    (Bottom) Binary foot contact states.
    }

    \label{fig:transition_wave}
\end{figure}

We demonstrate in simulation that a unified policy transitions from land to water, shifting from limb-driven walking to body-driven undulatory swimming, as visualized in Fig.~\ref{fig:transition_snapshots}.
We train the policy in the amphibious environment shown in Fig.~\ref{fig:transition_env}, using the observation formulation described in Sect.~\ref{transition_formulation}, which includes the binary mode indicator $\sigma$ in Eq.~\ref{transition_obs}. All training hyperparameters follow Tab.~\ref{tab:ppo_config} without additional fine-tuning.
To evaluate the transition, we initialize the robot on land near the water boundary, facing toward the fluid region.
A constant forward velocity command is applied, driving the robot to move from land into water under a single policy.
The resulting behavior is illustrated in Fig.~\ref{fig:transition_wave}.
The emergent spine and tail traveling waves during this transition qualitatively resemble the land–water transition waves observed in biological salamanders~\cite{IJSPEERT2008642}.
The top panel shows body joint trajectories over time. As the robot transitions toward swimming, the spine and tail joints develop traveling wave patterns.
The middle panel presents leg joint trajectories for all four limbs, where the limb joints gradually decrease as propulsion shifts from leg stepping to body undulation.
The bottom panel shows binary foot contact states, where contacts progressively disappear, marking the transition from walking to swimming.
Together, these results demonstrate that a single unified policy enables seamless transition from terrestrial walking to aquatic propulsion without controller switching.

\section{Conclusion}

This work presents an RL framework for joint-level whole-body control of a highly articulated salamander robot. The learned policy directly commands all actuated joints and achieves natural terrestrial locomotion with successful sim-to-real transfer on hardware.
Beyond terrestrial walking, we extend the framework to amphibious environments and demonstrate that a single unified policy transitions from land to water without controller switching. 
Future work will focus on enabling seamless walking and swimming transitions on physical hardware.

\section*{Acknowledgment}
We used LLMs (e.g., ChatGPT) to assist with language editing and manuscript polishing during the preparation of this paper.

\bibliography{refs}

\bibliographystyle{IEEEtran}

\vfill

\end{document}